\documentclass{article}

\usepackage{microtype}
\usepackage{graphicx}
\usepackage{subfigure}
\usepackage{booktabs} 
\usepackage{multirow}

\usepackage[accepted]{icml2024}

\usepackage{amsmath}
\usepackage{amssymb}
\usepackage{mathtools}
\usepackage{amsthm}

\usepackage[capitalize,noabbrev]{cleveref}

\theoremstyle{plain}

\theoremstyle{definition}

\theoremstyle{remark}

\usepackage[textsize=tiny]{todonotes}

\icmltitlerunning{Robust Federated Finetuning of Foundation Models via Alternating Minimization of LoRA}

\begin{document}

\twocolumn[
\icmltitle{Robust Federated Finetuning of Foundation Models \\ via Alternating Minimization of LoRA}




\begin{icmlauthorlist}
\icmlauthor{Shuangyi Chen}{yyy}
\icmlauthor{Yue Ju}{xxx}
\icmlauthor{Hardik Dalal}{xxx}
\icmlauthor{Zhongwen Zhu}{xxx}
\icmlauthor{Ashish Khisti}{yyy}
\end{icmlauthorlist}

\icmlaffiliation{yyy}{ECE Department, University of Toronto, Toronto, Canada}
\icmlaffiliation{xxx}{Ericsson-GAIA Montréal, Canada}

\icmlcorrespondingauthor{Shuangyi Chen}{shuangyi.chen@mail.utoronto.ca}
\icmlcorrespondingauthor{Ashish Khisti}{akhisti@ece.utoronto.ca}

\icmlkeywords{Machine Learning, ICML}

\vskip 0.3in
]



\printAffiliationsAndNotice{}  

\begin{abstract}
Parameter-Efficient Fine-Tuning (PEFT) has risen as an innovative training strategy that updates only a select few model parameters, significantly lowering both computational and memory demands. PEFT also helps to decrease data transfer in federated learning settings, where communication depends on the size of updates. In this work, we explore the constraints of previous studies that integrate a well-known PEFT method named LoRA with federated fine-tuning, then introduce RoLoRA, a robust federated fine-tuning framework that utilizes an alternating minimization approach for LoRA, providing greater robustness against decreasing fine-tuning parameters and increasing data heterogeneity. Our results indicate that RoLoRA not only presents the communication benefits but also substantially enhances the robustness and effectiveness in multiple federated fine-tuning scenarios.
\end{abstract}

\section{Introduction}
\label{intro}
The recent emergence of foundation models in various applications significantly changes the field of machine learning. Characterized by their broad adaptability and massive scale, these models require access to vast and diverse datasets to effectively learn across different tasks and domains. However, this presents a significant challenge: foundation models not only require large amounts of data, but also data of high quality.
Federated learning provides a promising solution to this issue. It enables the use of data from multiple sources while protecting the privacy of the data. By combining insights from different decentralized sources, federated learning allows for collaborative model training without exposing sensitive information. This method is especially beneficial for foundation models, as it can access a broad range of data while maintaining privacy. 

Recently, Parameter-Efficient Fine-Tuning (PEFT) has emerged as an innovative training strategy that updates only a small subset of model parameters, substantially reducing computational and memory demands. A notable method in this category is LoRA \cite{hu2021lora}, which utilizes low-rank matrices to approximate weight changes during fine-tuning. These matrices are integrated with pre-trained weights for inference, facilitating reduced data transfer in scenarios such as federated learning, where update size directly impacts communication efficiency. Many works integrate LoRA into federated setting. For example, FedPETuning \cite{zhang-etal-2023-fedpetuning} compared various PEFT methods in a federated setting. SLoRA \cite{babakniya2023slora}, a hybrid approach that combines sparse fine-tuning with LoRA, is introduced to tackle data heterogeneity in federated settings. Furthermore, FS-LLM \cite{kuang2023federatedscopellm} is presented, which is a framework for fine-tuning LLMs in federated environments. However, these studies typically apply the FedAVG algorithm directly to LoRA modules, overlooking the interference introduced by this aggregation approach. With this consideration, Sun et al. designs a federated finetuning framework named FFA-LoRA \cite{sun2024improving} based on LoRA by freezing down-projection matrix $\mathbf{A}$ for all the clients and only updating up-projection matrix $\mathbf{B}$. Furthermore, they apply DP-SGD to preserve privacy. Using sufficient number of finetuning parameters, FFA-LoRA with a larger learning rate achieves performance comparable to FedAVG for LoRA modules while halving the communication costs. However, we observe that with fewer fine-tuning parameters, FFA-LoRA is less robust than FedAVG for LoRA modules, primarily due to its limited expressiveness stemming from the restricted number of trainable parameters. 
Another common issue in federated learning is data heterogeneity among clients. To address this, we drew inspiration from the personalized federated framework FedRep \cite{pmlr-v139-collins21a}, which alternates between updating clients' representation and head. This approach highlights the importance of learning a robust low-rank representation and demonstrates superior convergence speed compared to simultaneously updating both representation and head.
Therefore, we propose a robust federated fine-tuning framework, RoLoRA, based on alternating minimization of LoRA. Empirical evidence demonstrates that RoLoRA is more robust against \textbf{Decreasing Fine-tuning Parameters} and \textbf{Increasing Data Heterogeneity}, while still halving communication costs, similar to FFA-LoRA.
\paragraph{Related Work} We provide a summary of the literature on PEFT, Variants of LoRA, PEFT in Federated Setting, and FL with data heterogeneity in  Appendix~\ref{related-works}.

\section{Preliminaries}
\subsection{Low-Rank Adaptation: LoRA} \label{lora-federated}
Low-Rank Adaptation (LoRA) \cite{hu2021lora} fine-tunes large language models efficiently by maintaining the original model weights fixed and adding small, trainable matrices in each layer. These matrices perform low-rank decompositions of updates, reducing the number of trainable parameters. This approach is based on the finding that updates to model weights during task-specific tuning are usually of low rank, which allows for fewer parameters to be adjusted. For example, for a pre-trained weight matrix $\mathbf{W}_0 \in \mathbb{R}^{d\times d}$, the update is a low-rank product $\mathbf{B}\mathbf{A}$, where $\mathbf{A}\in \mathbb{R}^{r\times d}$ and $\mathbf{B} \in \mathbb{R}^{d\times r}$, with $r << d$. Only $\mathbf{A}$ and $\mathbf{B}$ are trainable, allowing $\mathbf{W} = \mathbf{W}_0 + \alpha \mathbf{B}\mathbf{A}$, with $\alpha$ adjusting the update's impact.
Applying LoRA in a federated setting is a practical choice. By using LoRA adapters, clients can fine-tune foundation models efficiently with limited resources. Since only these specific matrices need to be transmitted to a central server, this approach significantly reduces communication costs. This makes LoRA an advantageous solution for enhancing model performance in collaborative scenario comparing to full parameter finetuning in the federated setting. 
\subsection{FedAVG of LoRA Introduces Interference}\label{interference}
Integrating LoRA within a federated setting presents challenges. In such a setup, each of the $N$ clients is provided with the pretrained model weights $\mathbf{W}_0$, which remain fixed during finetuning. Clients are required only to send the updated matrices $\mathbf{B}_i$ and $\mathbf{A}_i$ to a central server for aggregation. While most current studies, such as SLoRA\cite{babakniya2023slora} and FedPETuning\cite{zhang-etal-2023-fedpetuning}, commonly apply FedAVG directly to these matrices as shown in \eqref{fedavg-lora}, this approach might not be optimal. The precise update for each client’s model, $\Delta \mathbf{W}_i$, should be calculated as the product of the low-rank matrices $\mathbf{A}_i$ and $\mathbf{B}_i$. Consequently, aggregation on the individual matrices introduces interference. 
{\footnotesize
\begin{align}
    \frac{1}{N} \sum_{i=1}^N  \Delta \mathbf{W}_i = \frac{1}{N} (\mathbf{B}_1\mathbf{A}_1+\mathbf{B}_2\mathbf{A}_2+...+ \mathbf{B}_N\mathbf{A}_N) \label{fedavg-mul}\\
    \neq  \frac{1}{N} (\mathbf{B_1}+\mathbf{B_2}+...+\mathbf{B_N}) \frac{1}{N} (\mathbf{A_1}+\mathbf{A_2}+...+\mathbf{A_N}) \label{fedavg-lora}
\end{align}}
\subsection{FedRep: Common Representation via Alternating Minimization} \label{fedrep}
A common challenge in federated learning is data heterogeneity among clients. FedRep \cite{pmlr-v139-collins21a} addresses this by finding a common representation, which is effectively achieved via alternatively updating clients' representation and head, aggregating only representation while keeping the head diverse. The algorithm demonstrates the necessity of learning a robust low-rank representation. Additionally, the alternating optimization has shown superior convergence speed compared to approaches that simultaneously update both representation and head. We observe a structured similarity between the LoRA adapter and the representation-head structure. Specifically, we consider the down-projection matrix $\mathbf{A}$ in each LoRA adapter as the low-rank representation for the features of the intermediate layers. We hypothesize that learning a robust low rank representation (down-projection matrix) is also advantageous for the intermediate features when the clients has heterogeneous inputs. However, since the LoRA adapters are cascaded in the model unlike single representation-head structure in model considered in FedRep, keeping up-projection matrix $\mathbf{B}$ diverse may not be favorable for convergence. 

With these considerations, we propose robust federated fine-tuning framework based on alternating minimization of LoRA (RoLoRA).


\
\section{Our Framework} \label{rolora-framework}
We describe the framework design of RoLoRA and discuss its practical advantages. 
\paragraph{Alternating Minimization and Corresponding Aggregation}
Motivated by the observations discussed in Section~\ref{interference} and \ref{fedrep}, we propose applying alternating minimization to the local fine-tuning of each client in a setting with $N$ clients. Unlike the approach in FFA-LoRA, where $\mathbf{A}$ is consistently frozen, we suggest a alternating update strategy. There are alternating odd and even communication rounds designated for updating, aggregating $\mathbf{A}$ and $\mathbf{B}$, respectively. {\footnotesize
\begin{align}
 &\text{In the odd comm. round:} \qquad  \frac{1}{N} \sum_{i=1}^N  \Delta \mathbf{W}_{i}^{2t+1} \nonumber\\&= \frac{1}{N} (\mathbf{B}_1^{t+1}\mathbf{A}_1^t+\mathbf{B}_2^{t+1}\mathbf{A}_2^t+...+ \mathbf{B}_N^{t+1}\mathbf{A}_N^t) \label{round-t}\\
  &= \frac{1}{N} (\mathbf{B}_1^{t+1}+\mathbf{B}_2^{t+1}+...+ \mathbf{B}_N^{t+1})\mathbf{A}^t  \nonumber\\
  & \text{In the even comm. round:} \quad  \frac{1}{N} \sum_{i=1}^N  \Delta \mathbf{W}_{i}^{2t+2} \nonumber \\&= \frac{1}{N} (\mathbf{B}_1^{t+1}\mathbf{A}_1^{t+1}+\mathbf{B}_2^{t+1}\mathbf{A}_2^{t+1}+...+ \mathbf{B}_N^{t+1}\mathbf{A}_N^{t+1}) \label{roundt+1}\\
&= \frac{1}{N} \mathbf{B}^{t+1}(\mathbf{A}_1^{t+1}+\mathbf{A}_2^{t+1}+...+ \mathbf{A}_N^{t+1}) \nonumber
\end{align}
} 
In the odd communication round, all clients freeze $\mathbf{A}^t$ and update $\mathbf{B}^{t}$. The central server then aggregates these updates to compute $\mathbf{B}^{t+1} = \frac{1}{N}\sum_{i=1}^{N}\mathbf{B}^{t+1}_i$ and distributes $\mathbf{B}^{t+1}$ back to the clients. In the subsequent communication round, clients freeze $\mathbf{B}^{t+1}$ and update $\mathbf{A}^{t}$. The server aggregates these to obtain $\mathbf{A}^{t+1} = \frac{1}{N}\sum_{i=1}^{N}\mathbf{A}^{t+1}_i$ and returns $\mathbf{A}^{t+1}$ to the clients. 
It is important to note that in round $2t+1$, the frozen $\mathbf{A}_i^t$ are identical across all clients, as they are synchronized with $\mathbf{A}^t$ from the central server at the beginning of the round. This strategy ensures that the update and aggregation method introduces no interference, as demonstrated in \eqref{round-t} and \eqref{roundt+1}.

\paragraph{Computation and Communication Cost}
The parameter-freezing nature of RoLoRA enhances computational and communication efficiency. In each communication round, the number of trainable parameters in the model is effectively halved compared to FedAVG with LoRA. The only additional cost for RoLoRA compared to FFA-LoRA is the alternating freezing of the corresponding parameters. We remark this additional cost is negligible because it is applied to the clients' models and can be executed concurrently during the server's aggregation.
\section{Experiments}\label{exp}
We evaluate the performance of RoLoRA in various federated settings. We use NVIDIA GeForce RTX 4090 or NVIDIA A40 for all the experiments.
\begin{figure*}[ht]
\begin{center}
\begin{subfigure}
  \centering
  \includegraphics[width=0.185\linewidth]{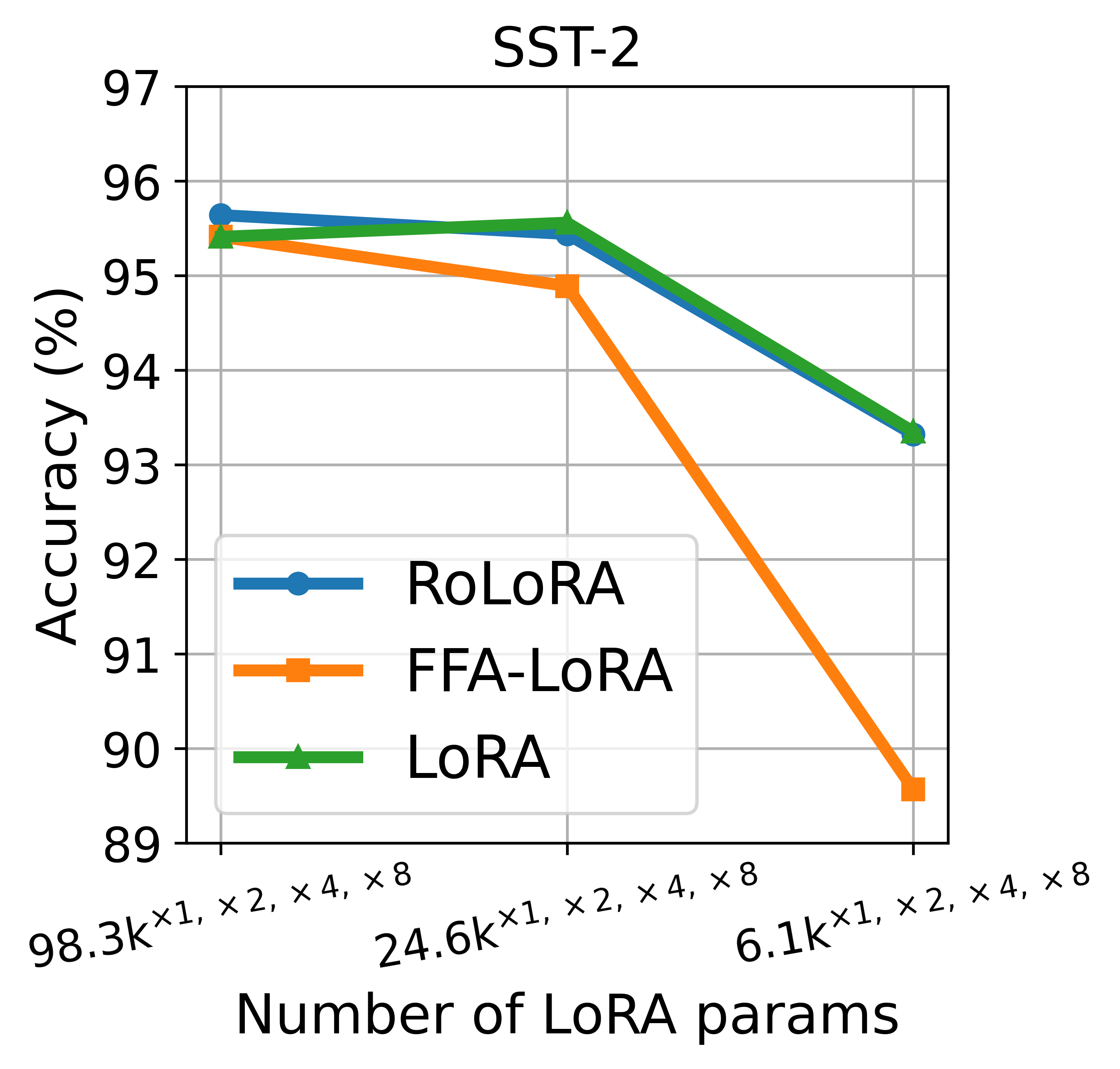}
  \label{fig:sub1}
\end{subfigure}
\hfill
\begin{subfigure}
  \centering
  \includegraphics[width=0.185\linewidth]{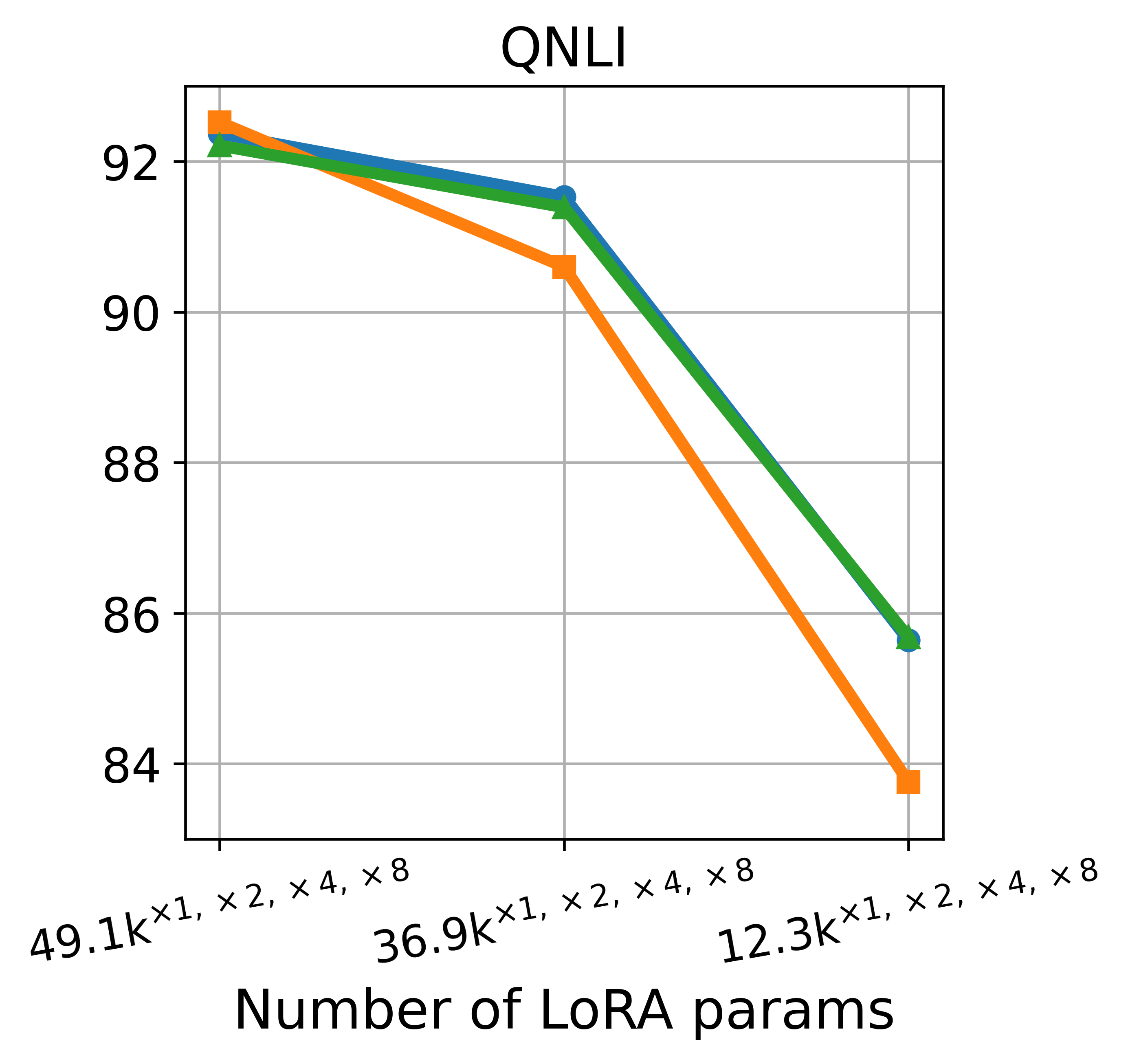}
  \label{fig:sub2}
\end{subfigure}
\hfill
\begin{subfigure}
  \centering
  \includegraphics[width=0.185\linewidth]{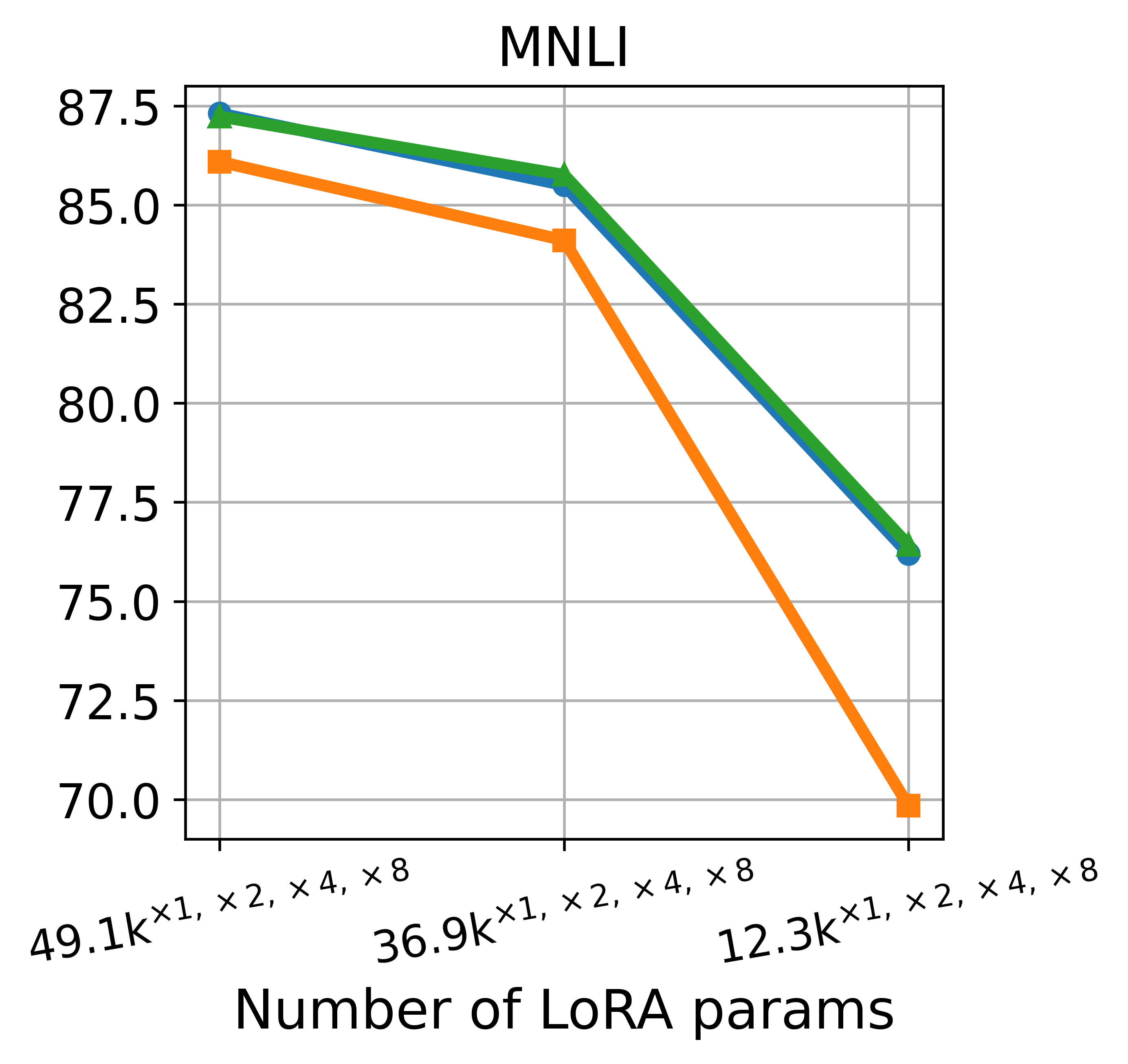}
  \label{fig:sub3}
\end{subfigure}
\hfill
\begin{subfigure}
  \centering
  \includegraphics[width=0.185\linewidth]{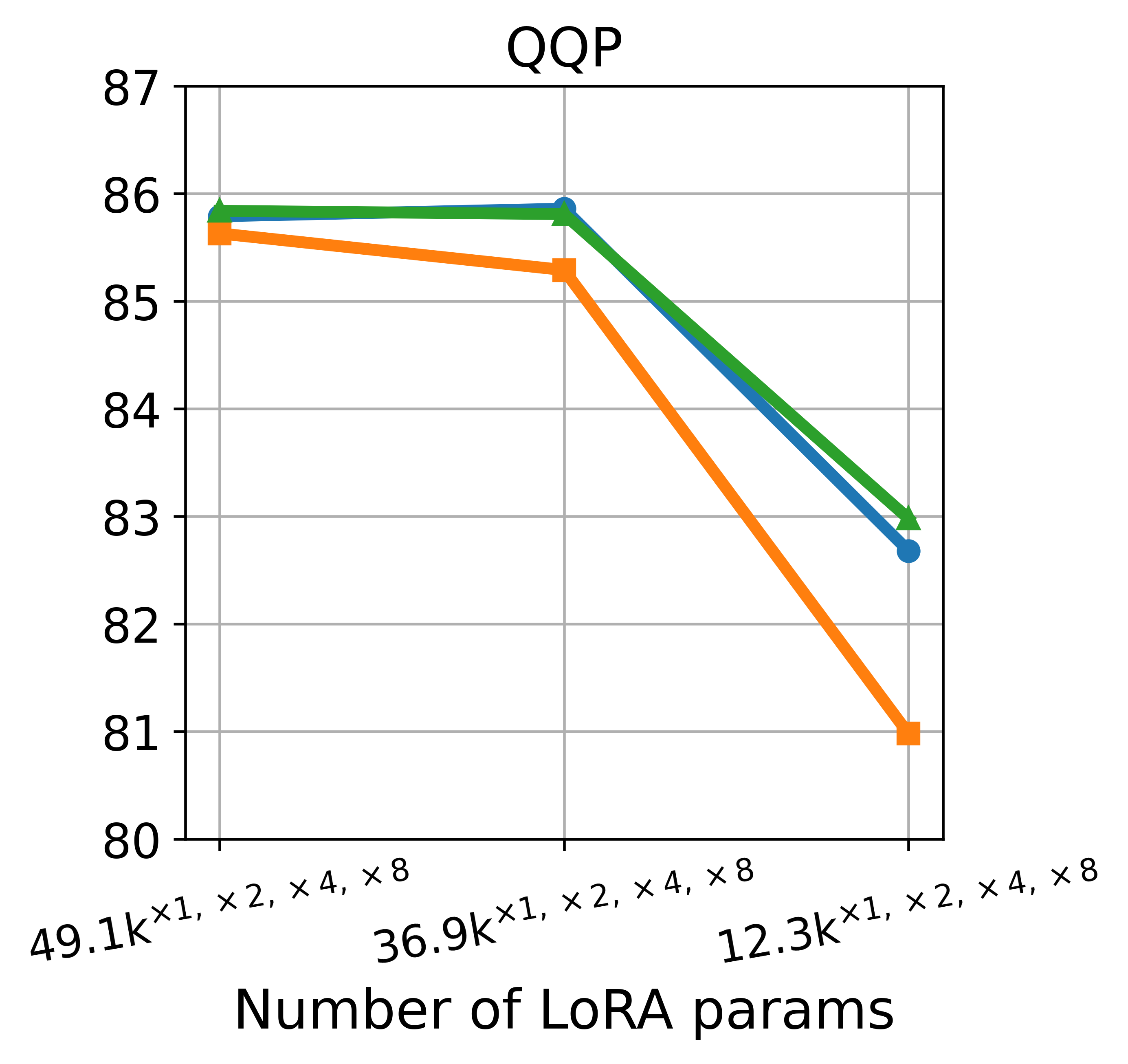}
  \label{fig:sub4}
\end{subfigure}
\hfill
\begin{subfigure}
  \centering
  \includegraphics[width=0.185\linewidth]{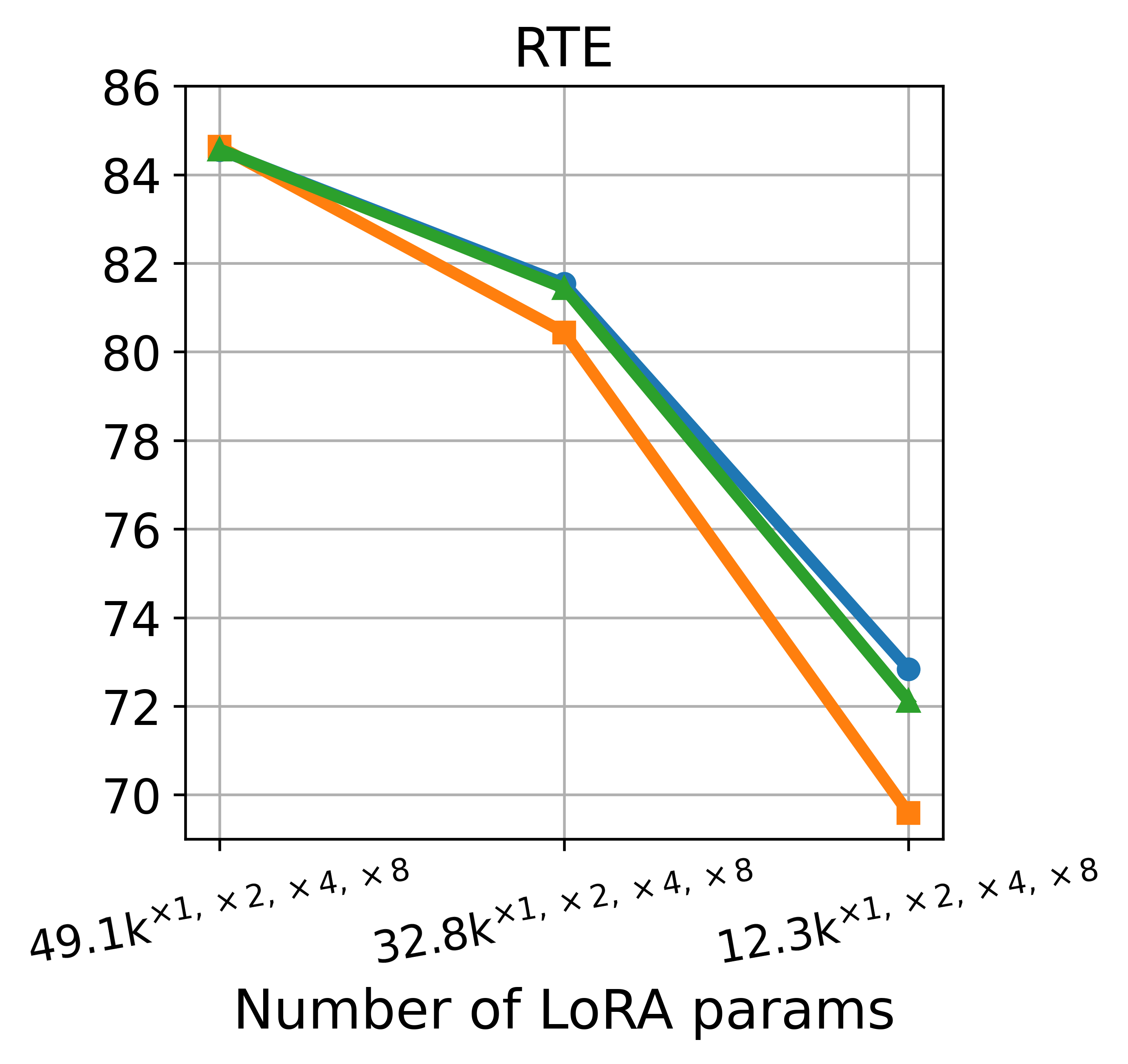}
  \label{fig:sub5}
\end{subfigure}
 \caption{Results with RoBERTa-Large models on GLUE of different budget of finetuning parameters. The accuracy is computed by averaging over different ranks $\{1,2,4,8\}$. The number of clients is 3.}
    \label{fig:five_subfigures}
\hfill
\end{center}
\end{figure*}
\begin{table*}
{\scriptsize
    \centering
    \begin{tabular}{ccccccccc}
    \toprule
 Rank  &Methods &Comm. cost    & SST-2 & QNLI & MNLI & QQP  & RTE & Avg.\\
      \midrule 
  &LoRA   & $\times 16$ & $\text{95.68}_{\pm \text{0.14}}$ & $\text{91.46}_{\pm \text{0.30}}$& $\text{85.93}_{\pm \text{0.01}}$  & $\text{85.95}_{\pm \text{0.18}}$ & $\text{81.35}_{\pm \text{0.74}}$  &88.07\\
r=8  &FFA-LoRA & $\times 8$  & $\text{94.99}_{\pm \text{0.10}}$ & $\text{91.09}_{\pm \text{0.36}}$ &$\text{85.21}_{\pm \text{0.03}}$  &$\text{85.76}_{\pm \text{0.08}}$  & $\text{80.14}_{\pm \text{1.02}}$ & 87.44\\
  &RoLoRA & $\times 8$  & $\text{95.45}_{\pm \text{0.14}}$ &$\text{91.84}_{\pm \text{0.09}}$  & $\text{85.76}_{\pm \text{0.01}}$ &$\text{85.91}_{\pm \text{0.22}}$  & $\text{81.32}_{\pm \text{0.78}}$ &88.06\\
  \midrule 
  &LoRA  & $\times 8$  & $\text{95.62}_{\pm \text{0.17}}$ & $\text{91.59}_{\pm\text{0.21}}$ & $\text{86.20}_{\pm \text{0.05}}$ & $\text{86.13}_{\pm \text{0.10}}$ & $\text{81.46}_{\pm \text{1.22}}$ & 88.20\\
r=4  &FFA-LoRA  & $\times 4$ & $\text{95.18}_{\pm \text{0.09}}$ & $\text{91.35}_{\pm \text{0.32}}$ & $\text{84.58}_{\pm \text{0.21}}$ & $\text{85.50}_{\pm \text{0.25}}$ & $\text{81.10}_{\pm \text{0.33}}$ & 87.48\\
  &RoLoRA & $\times 4$ & $\text{95.49}_{\pm \text{0.16}}$ & $\text{91.64}_{\pm \text{0.30}}$ & $\text{85.70}_{\pm \text{0.04}}$ & $\text{86.14}_{\pm \text{0.06}}$ & $\text{82.43}_{\pm \text{0.84}}$ & 88.28\\
  \midrule 
  &LoRA &$\times 4$  & $\text{95.64}_{\pm \text{0.11}}$ & $\text{92.04}_{\pm \text{0.11}}$ & $\text{85.85}_{\pm \text{0.19}}$ & $\text{86.16}_{\pm \text{0.08}}$ & $\text{82.19}_{\pm \text{1.03}}$ & 88.38\\
r=2   &FFA-LoRA&$\times 2$   & $\text{94.91}_{\pm \text{0.16}}$ & $\text{90.11}_{\pm \text{0.17}}$ & $\text{84.06}_{\pm \text{0.19}}$ & $\text{85.48}_{\pm \text{0.01}}$ & $\text{80.86}_{\pm \text{0.51}}$ & 87.08\\
  &RoLoRA  & $\times 2$  & $\text{95.60}_{\pm \text{0.10}}$ & $\text{91.62}_{\pm \text{0.32}}$ & $\text{85.55}_{\pm \text{0.05}}$ & $\text{86.16}_{\pm \text{0.18}}$ & $\text{82.19}_{\pm \text{1.03}}$ & 88.22\\
  \midrule 
  &LoRA &$\times 2$ & $\text{95.32}_{\pm \text{0.18}}$ & $\text{90.48}_{\pm \text{0.56}}$ & $\text{85.08}_{\pm \text{0.04}}$ & $\text{85.01}_{\pm \text{0.05}}$ & $\text{81.10}_{\pm \text{0.95}}$&87.40\\
r=1   &FFA-LoRA&$\times 1$   & $\text{94.49}_{\pm \text{0.22}}$ & $\text{89.87}_{\pm \text{0.37}}$ & $\text{82.60}_{\pm \text{0.03}}$ & $\text{84.42}_{\pm \text{0.50}}$ & $\text{79.66}_{\pm \text{1.08}}$ &86.21\\
  &RoLoRA & $\times 1$ & $\text{95.22}_{\pm \text{0.14}}$ & $\text{91.01}_{\pm \text{0.23}}$ & $\text{84.97}_{\pm \text{0.05}}$ & $\text{85.24}_{\pm \text{0.18}}$ & $\text{80.23}_{\pm \text{1.02}}$ &87.33\\
  \bottomrule
    \end{tabular}
    \caption{Results with RoBERTa-Large models on GLUE. We report the average and std. over five seeds. The number of clients is 3. Please refer to Table~\ref{tab:comm_size} in Appendix~\ref{ft-dynamics} for the actual communication cost.}
    \label{tab:glue1}}
\end{table*}
\paragraph{Baselines} Considering cross-silo federated setting where the number of clients is relatively small and all clients will participate in each round, we will explore the following three methods based on FedAVG.
\begin{itemize}
    \item \textbf{LoRA} means LoRA adapter and its finetuning algorithm are directly applied to local finetuning of clients in the federated system. Specifically, in iteration $t$, the server receives $\mathbf{A}_i^{t}$ and $\mathbf{B}_i^{t}$ from client $i$ and aggregates by  $\mathbf{A}^{t} = \mathsf{Avg}(\mathbf{A}_i^{t})$ and $\mathbf{B}^{t} = \mathsf{Avg}(\mathbf{B}_i^{t})$. 
    \item \textbf{LoRA-FFA} \cite{sun2024improving} is a baseline that enable the clients to finetune $\mathbf{B}$ and keep $\mathbf{A}$ frozen locally. Thus, in iteration $t$, the server aggregates by $\mathbf{B}^{t} = \mathsf{Avg}(\mathbf{B}_i^{t})$. 
    \item \textbf{RoLoRA} enables clients to alternate updating $\mathbf{A}$ and $\mathbf{B}$ as described in Section~\ref{rolora-framework}.  
\end{itemize}

\paragraph{Model and Datasets.} We take the pre-trained RoBERTa-Large (355M) \cite{liu2019roberta} models from the HuggingFace Transformers library. and evaluate the performance of three federated finetuning methods on 5 datasets (SST-2, QNLI, MNLI, QQP, RTE) from the GLUE \cite{wang2019glue}. Due to the limitation of the unpublished test set in GLUE, we follow the previous studies \cite{zhang-etal-2023-fedpetuning} and use the original validation set as the new test set and split a part of the training set as the validation set. 

\paragraph{Implementation.}We implement all the methods based on FederatedScope-LLM \cite{kuang2023federatedscopellm}. To make a fair comparison, for each dataset, we obtain the best performance on test set and report the average over five seeds. Specifically, the learning rate is chosen from the set $\{5e-4, 1e-3, 2e-3, 5e-3, 1e-2, 2e-2, 5e-2, 1e-1, 2e-1\}$. Other hyper-parameters for experiments are specified in Table~\ref{tab:exp-set} in Appendix~\ref{exp-setup}.

\paragraph{Effect of Number of Finetuning Parameters}
In Figure~\ref{fig:five_subfigures}, we compare three methods across five GLUE datasets. We apply LoRA to every weight matrix of the selected layers, given different budgets of LoRA parameters. For each dataset, we experiment with three budgets, ranging from high to low. The corresponding layer sets, ${\mathcal{P}_1, \mathcal{P}_2, \mathcal{P}_3}$, are detailed in Table~\ref{tab:layer_type_index} in Appendix~\ref{exp-setup}.

The figures indicates that with sufficient number of finetuning parameters, the three methods can achieve comparable best accuracy; as the number of LoRA parameters is reduced, the performance of the three methods deteriorates to varying degrees. However, RoLoRA, which achieves performance comparable to LoRA, demonstrates greater robustness compared to FFA-LoRA, especially under conditions of limited fine-tuning parameters. It is important to note that with the same finetuning parameters, the communication cost of RoLoRA and FFA-LoRA is always half of that of LoRA due to their parameter freezing nature. This implies that RoLoRA not only sustains its performance but also enhances communication efficiency. We expand the middle set of data of each of Figure~\ref{fig:five_subfigures}, corresponding to $\mathcal{P}_2$, and show the details of the performance of three methods in Table~\ref{tab:glue1}. 

\begin{table}[t]
{\scriptsize
\centering
    \begin{tabular}{ccccccc}
    \toprule
        &  & SST-2 & QNLI & MNLI & QQP & RTE \\
        \midrule 
         &LoRA&{95.64}&{92.04}&{85.85}&{86.16}&{82.19}\\      
        iid. &FFA-LoRA&94.91&90.11&84.06&85.48&80.86\\
         &RoloRA&95.60&91.62&85.66&{86.16}&{82.19}\\
         \midrule 
                 &LoRA&94.27&86.91&81.22&82.07&46.21\\
        mild het. &FFA-LoRA&93.92&89.58&80.51&82.62&57.76\\
         &RoloRA&{94.84}&{90.77}&{85.13}&{85.10}&{81.23}\\
         \midrule 
                 &LoRA&93.23&82.57&58.96&76.96&49.10\\
        severe het. &FFA-LoRA&92.32&85.15&62.79&77.78&53.07\\
         &RoloRA&{94.61}&{89.83}&{85.15}&{85.55}&{72.92} \\
         \bottomrule
    \end{tabular}
        \caption{Results with RoBERTa-Large models with varying client numbers (3, 20, 50) to increase data heterogeneity in federated setting, maintaining a constant sample count during fine-tuning. The rank used is 2.
        \label{tab:Clients-num}}
}
\end{table}

\paragraph{Effect of Data Heterogeneity}
In this section, we study the effect of data heterogeneity. The layer set with LoRA adapters in Table~\ref{tab:Clients-num} is $\mathcal{P}_2$ as in Table~\ref{tab:glue1}. In Table~\ref{tab:Clients-num}, we increased the number of clients from 3 to 20, and then to 50, ensuring that there is no overlap in the training samples each client can access. Consequently, each client receives a smaller fraction of the total dataset, leading to a rise in data heterogeneity among the clients. We observe that as the data heterogeneity increases, while maintaining the same number of fine-tuning samples, the performance of the LoRA method significantly deteriorates for most datasets. In contrast, RoLoRA maintains its accuracy levels. The performance of FFA-LoRA also declines, attributed to the limited expressiveness of the random initialization of $\mathbf{A}$ for clients' heterogeneous data. Notably, RoLoRA achieves this accuracy while incurring only half the communication costs associated with LoRA. Figure~\ref{fig:convergence-speed} in Appendix~\ref{ft-dynamics} illustrates the dynamics during fine-tuning for three methods, highlighting that the convergence speed of RoLoRA is substantially better than that of the other two methods.

\paragraph{Align Communication Cost for Three Methods}

In Table~\ref{tab:same_comm}, we conduct a comparison of three methods under the constraint of identical communication costs under the assumption that the number of clients is small. To align the communication costs across these methods, two approaches are considered. The first approach involves doubling the rank of FFA-LoRA and RoLoRA, with the outcomes detailed in Table~\ref{tab:glue1}. The second approach requires doubling the number of layers equipped with LoRA adapters. In the results presented in Table~\ref{tab:same_comm}, the latter strategy is employed.  Specifically, for both FFA-LoRA and RoLoRA, we adjust the communication costs by doubling the number of layers equipped with LoRA adapters, compared to the baseline LoRA method, where the layer set $\mathcal{P}_3$ are attached with adapters, thereby standardizing the size of the transmitted messages. Table~\ref{tab:same_comm} demonstrates that when operating within a constrained communication cost budget, the performance of RoLoRA consistently surpasses that of the other two methods.
\begin{table}[t]
{\scriptsize
    \centering
    \begin{tabular}{ccccccc}
    \toprule
        Model/Rank & Methods  & QNLI & MNLI & QQP & RTE \\
        \midrule
         &LoRA   & 86.07 & 76.58 & 83.77& 73.58 \\
       $\textsf{Rob-L}_{r=4}$  &FFA-LoRA    & 88.54 & 78.28 & 84.04 & 74.64 \\
         &RoLoRA   & {89.13} & {82.33} &{84.58}  & {76.51}  \\
         \midrule
         & LoRA  & {85.47} & {76.26} & {82.32} &  69.68\\
       $\textsf{Rob-L}_{r=2}$  & FFA-LoRA  & 87.81 & 77.24 & 83.81 &  76.11 \\
         & RoLoRA   & {89.19} & {82.18} &{84.24}  & {76.53}  \\

           \bottomrule
    \end{tabular}
    \caption{Results with RoBERTa-Large models on GLUE. Same message size in each round for three methods for each dataset. The number of clients is 3.}
    \label{tab:same_comm}}
\end{table}

More experimental results with different models and settings are provided in Appendix~\ref{more-results}.
\section{Conclusion}
In this work, we introduce RoLoRA, a robust federated fine-tuning framework using alternating minimization for LoRA. RoLoRA improves robustness against reduced fine-tuning parameters and increased data heterogeneity. Our results show that RoLoRA enhances communication efficiency, robustness, and effectiveness in various federated fine-tuning settings.
\bibliography{example_paper}
\bibliographystyle{icml2024}

\newpage
\appendix
\onecolumn
\section{Related Works} \label{related-works}
\subsection{Parameter Efficient Fine Tuning (PEFT): LoRA and Its Variants}
As the size of large language models (LLMs) continues to increase, it is computationally expensive and time-consuming to finetune to the full model. Parameter efficient finetuning (PEFT) allows for updates to a smaller subset of parameters, significantly reducing the computational and memory requirements. One of the most well-known methods is LoRA\cite{hu2021lora}.
LoRA uses low-rank matrices to approximate changes in weights during fine-tuning, allowing them to be integrated with pre-trained weights before inference. Based on LoRA, many PEFT methods are developed. For example, Zhang et al. \cite{zhang2023adalora} designs AdaLoRA by using SVD decomposition and pruning less significant singular values for more efficient updates. VeRA \cite{kopiczko2024vera} is proposed to further reduce the number of trainable parameters during finetuning by using a single pair of low-rank matrices shared across all layers and learning small scaling vectors. Zhang et al. \cite{zhang2023lorafa} proposes a memory-efficient fine-tuning method named LoRA-FA which keeps the projection-down weight of $\mathbf{A}$ fixed and updates the projection-up weight of $\mathbf{B}$ during finetuning. Hayou et al. \cite{hayou2024lora} enhance LoRA by assigning different learning rates to $\mathbf{A}$ and $\mathbf{B}$, theoretically confirming that the optimal approach requires a higher learning rate for $\mathbf{B}$ than for $\mathbf{A}$. Liu et al. analyze magnitude and directional updates in LoRA versus full parameter fine-tuning and introduce DoRA\cite{liu2024dora}, which decomposes pre-trained weights for fine-tuning and applies LoRA for directional updates. A quantized version of LoRA named QLoRA\cite{dettmers2023qlora} is introduced. Building upon that, Li et al. develops LoftQ \cite{li2023loftq} for a better initialization for quantized training.
\subsection{PEFT in Federated Setting}

PEFT adjusts only a few lightweight or a small portion of the total parameters for specific tasks, keeping most foundational model parameters unchanged. This feature can help reduce data transfer in federated learning, where communication depends on the size of updates. Zhang et al. \cite{zhang-etal-2023-fedpetuning} compares multiple PEFT methods in federated setting, including Adapter\cite{houlsby2019parameterefficient}, LoRA\cite{hu2021lora}, Prompt tuning\cite{liu2022ptuning} and Bit-Fit\cite{zaken2022bitfit}. SLoRA\cite{babakniya2023slora}, which combines sparse finetuning and LoRA, is proposed by Babakniya et al. to address the data heterogeneity in federated setting. 
Sun et al. designs a federated finetuning framework named FFA-LoRA based on LoRA \cite{sun2024improving} by freezing matrix $\mathbf{A}$ for all the clients and only updating matrix $\mathbf{B}$. Furthermore, they apply DP-SGD to preserve privacy. FS-LLM \cite{kuang2023federatedscopellm}, a framework for finetuning LLM, is introduced.
\subsection{FL with Data Heterogeneity}
\paragraph{FL with a Common Representation}
FL with a common representation aims to address the challenges of data heterogeneity in FL. Those FL methods learn a shared global representation while allowing each client to have its own personalized partial model. Works include FedRep\cite{pmlr-v139-collins21a}, which learns a global low-dimensional representation and personalized head for each client, FedCR \cite{pmlr-v202-zhang23w}, which introduces a regularizer to encourage learning a shared representation, and FedPAC \cite{xu2023personalized}, which performs class-wise feature alignment. Other methods like FedBABU \cite{oh2022fedbabu} and FedRoD \cite{chen2022bridging} also aim to learn a shared representation across clients. Although we focus on a common model in the federated setting in this work, we got inspired by training algorithm introduced by FedRep \cite{pmlr-v139-collins21a} to learn a low-rank representation for the intermediate features. We discuss the similarity and difference between the LoRA adapter and representation-head structure. 

\section{Experiments} 
\subsection{Setup} \label{exp-setup}
We show the hyper-parameter configurations for each dataset in Table~\ref{tab:exp-set}.
\begin{table}[h]
    \centering
{\footnotesize
    \begin{tabular}{cccccc}
    \toprule
         & SST-2 & QNLI & MNLI & QQP & RTE \\
         \midrule
    Total comm. rounds     & 500&500&500&500&200 \\
    Batch Size &64&32&32&32&32\\
    Local Epochs & 20&20&20&20&20 \\
    \bottomrule
    \end{tabular}
    \caption{Hyper-parameters configurations}
    \label{tab:exp-set}}
\end{table}
\begin{table}[h]
\centering
{\footnotesize
    \begin{tabular}{ccccccc}
    \toprule
                            & Layer Attributes & SST-2 & QNLI & MNLI & QQP & RTE \\
                            \midrule
\multirow{2}{*}{$\mathcal{P}_1$} & Type             &  $W_v,W_q$     &   $W_v,W_q$    &   $W_v,W_q$    &  $W_v,W_q$    &  $W_v,W_q$    \\
                    & Index            &    $\{0,\ldots,23\}$  &  $\{12,\ldots,23\}$    &   $\{12,\ldots,23\}$   &   $\{12,\ldots,23\}$  &   $\{12,\ldots,23\}$  \\
\multirow{2}{*}{$\mathcal{P}_2$} & Type             &     $W_v,W_q$   &  $W_v,W_q$     &   $W_v,W_q$    &   $W_v,W_q$   &   $W_v,W_q$   \\
                    & Index            &   $\{18,\ldots,23\}$    &   $\{15,\ldots,23\}$   &  $\{15,\ldots,23\}$    &   $\{15,\ldots,23\}$  &  $\{16,\ldots,23\}$   \\
\multirow{2}{*}{$\mathcal{P}_3$} & Type             &   $W_v$    &    $W_v,W_q$    &   $W_v,W_q$     &    $W_v,W_q$   &    $W_v,W_q$   \\
                    & Index            &  $\{21,\ldots,23\}$  &   $\{21,\ldots,23\}$    &   $\{21,\ldots,23\}$    &   $\{21,\ldots,23\}$   &  $\{21,\ldots,23\}$   \\
                    \bottomrule
    \end{tabular}}
    \caption{The selected layer set attached with LoRA}
    \label{tab:layer_type_index}
\end{table}

In Table~\ref{tab:layer_type_index}, we include the details about layers attached with LoRA adapters for different budget of finetuning parameters, for each dataset.
\subsection{More Results} \label{more-results}
 
\subsubsection{Communication Cost}
In Table~\ref{tab:comm_size}, we show the uplink communication cost for three methods for the layer set $\mathcal{P}_2$ using rank=1.
\begin{table}[h]
{\footnotesize
    \centering
    \begin{tabular}{ccccccc}
    \toprule
 Rank  &Methods     & SST-2 & QNLI & MNLI & QQP  & RTE \\
      \midrule 
  &LoRA     &46.9 & 93.8 & 93.8& 140.6 & 125  \\
r=1  &FFA-LoRA    & 23.5 & 46.9 & 46.9& 70.3 & 62.5\\
  &RoLoRA    & 23.5 & 46.9 & 46.9 & 70.3  & 62.5 \\

  \bottomrule
    \end{tabular}
    \caption{Uplink message size (KB) in each communication round for experiments in Table~\ref{tab:glue1} when rank=1. The message size will proportionally increase if utilizing different ranks.}
    \label{tab:comm_size}}
\end{table}

\subsubsection{Finetuning Dynamics of the Setup with Severe Data Heterogeneity} \label{ft-dynamics}
\begin{figure*}[h]
\begin{center}
\begin{subfigure}
  \centering
  \includegraphics[width=0.185\linewidth]{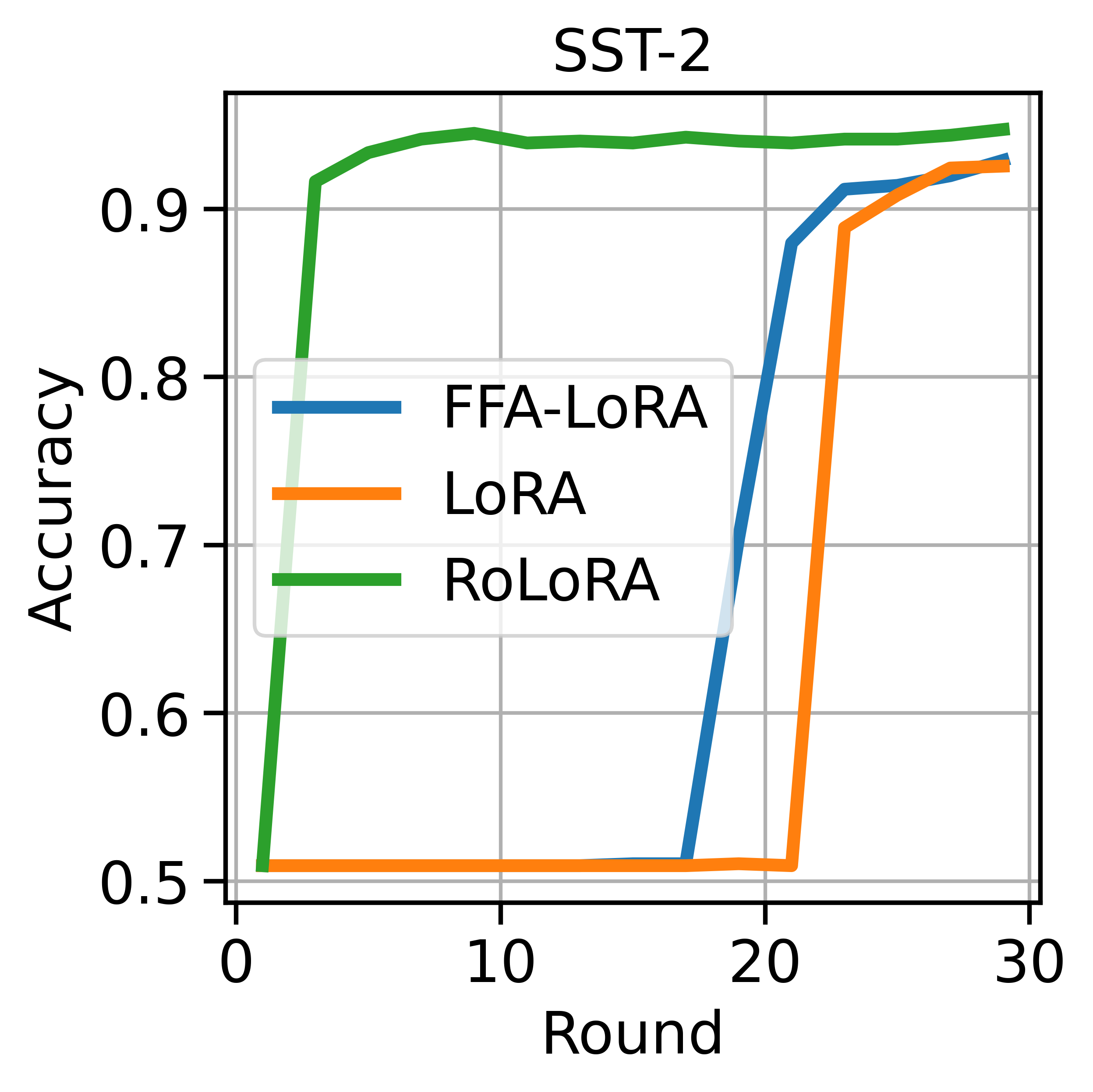}
  \label{fig:sub1-}
\end{subfigure}
\hfill
\begin{subfigure}
  \centering
  \includegraphics[width=0.18\linewidth]{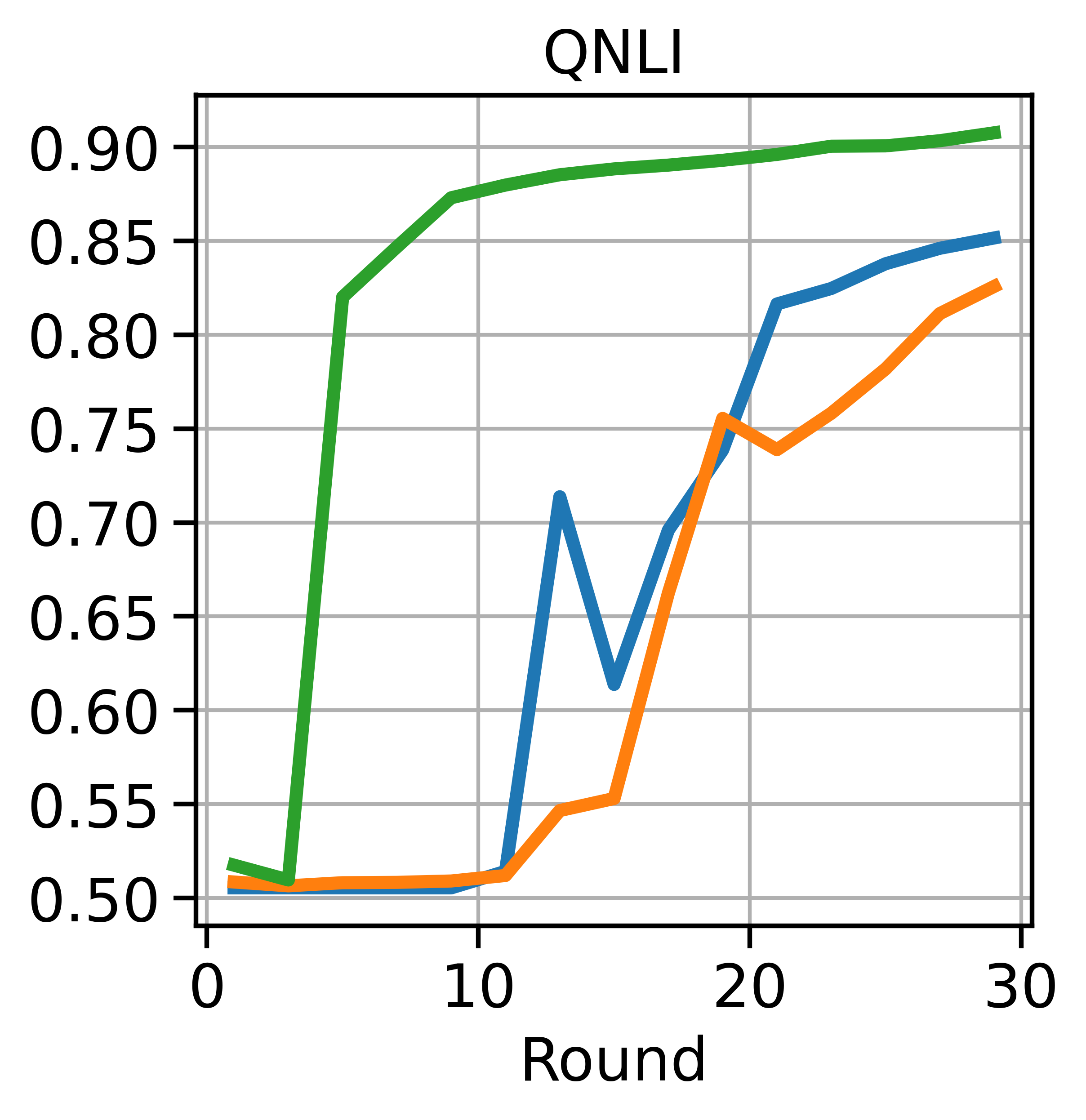}
  \label{fig:sub2-}
\end{subfigure}
\hfill
\begin{subfigure}
  \centering
  \includegraphics[width=0.174\linewidth]{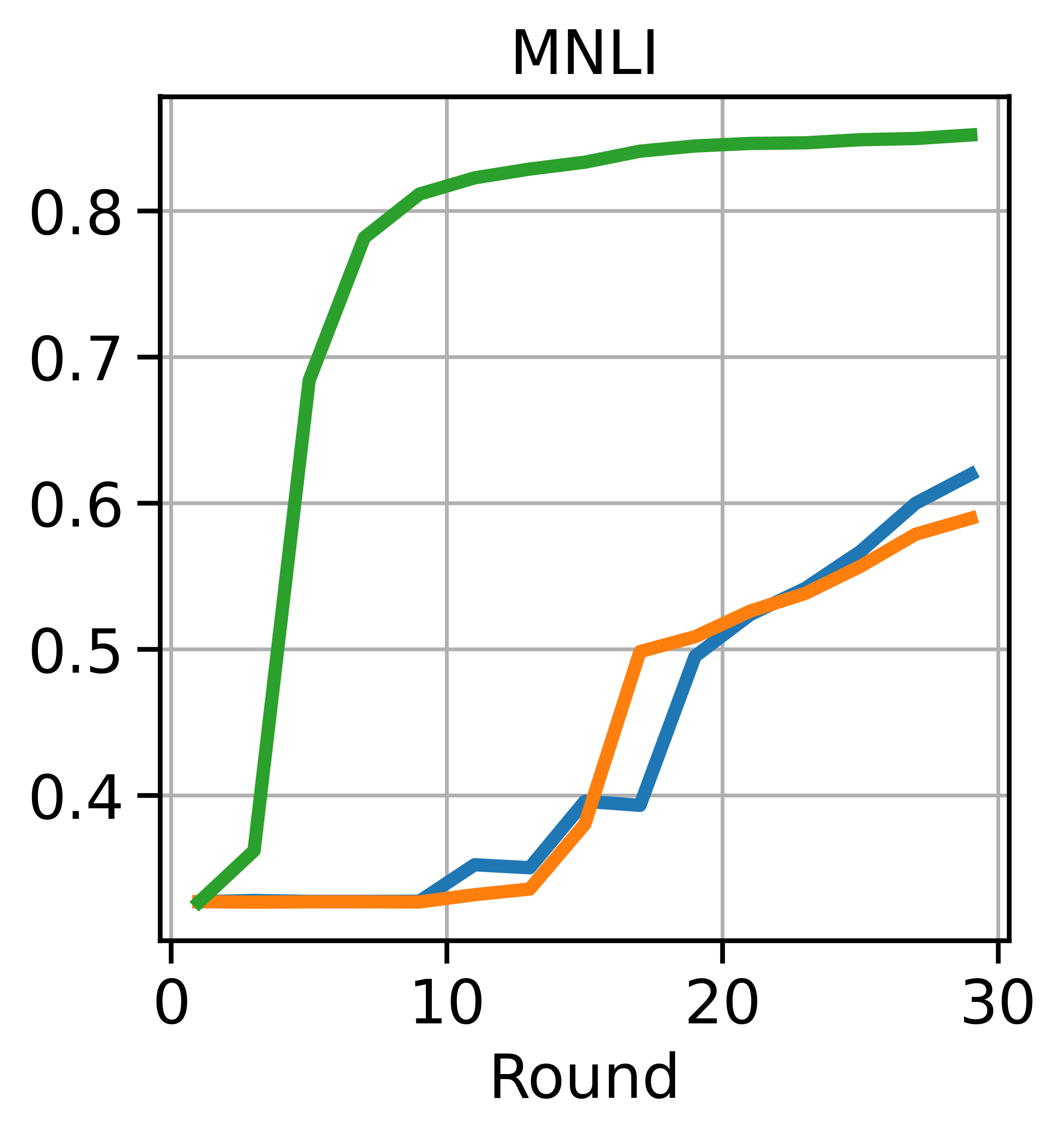}
  \label{fig:sub3-}
\end{subfigure}
\hfill
\begin{subfigure}
  \centering
  \includegraphics[width=0.181\linewidth]{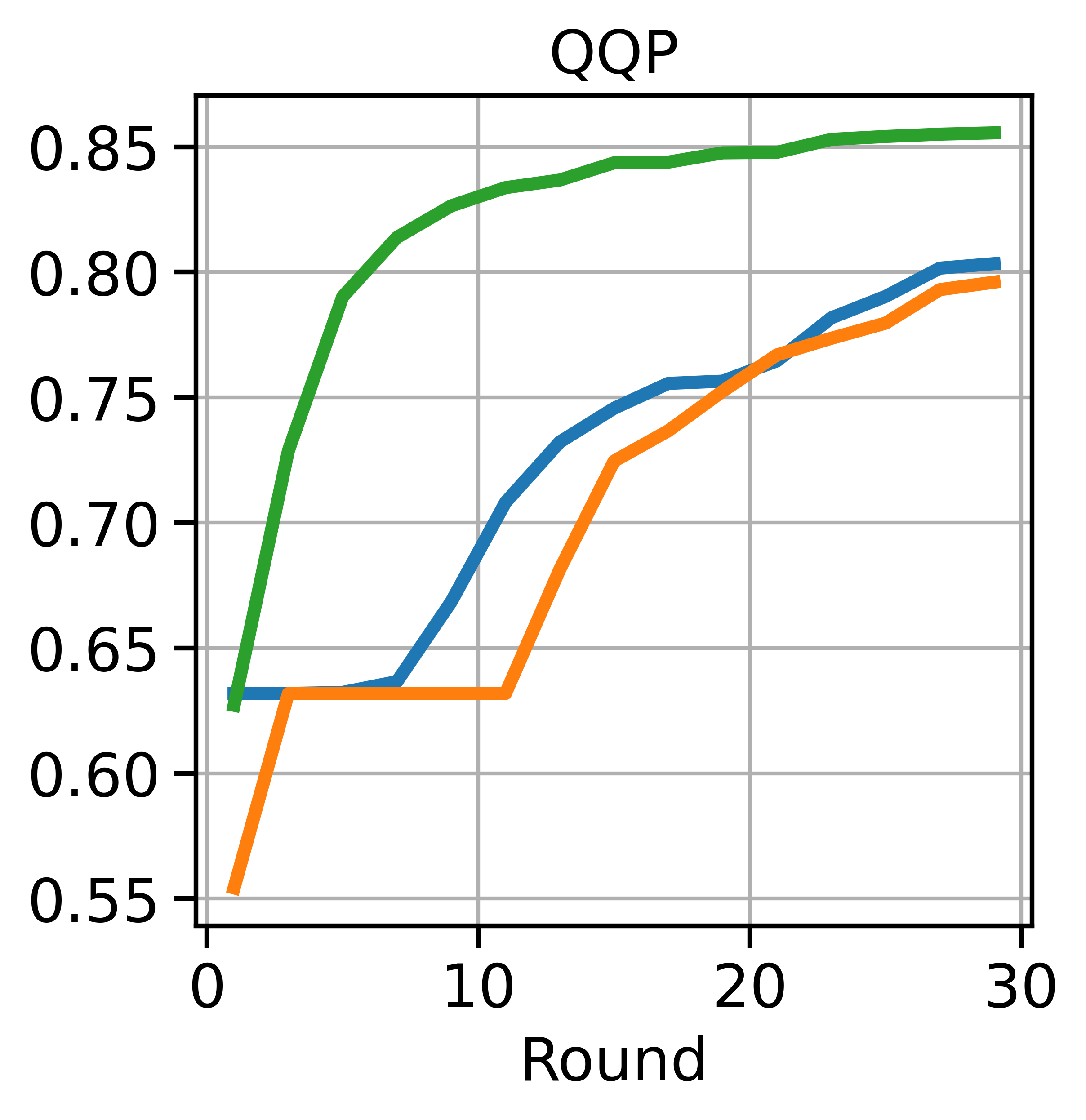}
  \label{fig:sub4-}
\end{subfigure}
 \caption{Accuracies over rounds with RoBERTa-Large models on SST-2, QNLI, MNLI, and QQP. The total number of clients is 50.}
    \label{fig:convergence-speed}
\hfill
\end{center}
\end{figure*}

In Figure~\ref{fig:convergence-speed}, we show the convergence of three methods under severe data heterogeneity with 50 clients. RoLoRA demonstrates superior convergence speed compared to the other two methods.

\newpage
\subsubsection{DeBERTa-XLarge Results} 
In Table~\ref{tab:same_comm_deb}, we show the results with DeBERTa-XLarge (900M). For both FFA-LoRA and RoLoRA, we modify the communication costs by equipping twice as many layers with LoRA adapters compared to the standard LoRA method. So the communication costs are aligned for three methods.
\begin{table}[h]
{\footnotesize
    \centering
    \begin{tabular}{ccccccc}
    \toprule
        Model/Rank & Methods  & QNLI & MNLI & QQP & RTE \\
        \midrule
        &LoRA   & 90.16 & 83.58 & 85.67 & 74.97 \\
       $\textsf{Deb-XL}_{r=4}$  &FFA-LoRA    & 90.08 & 83.31 & 85.79 &  78.73 \\
         &RoLoRA   & 91.36 & 84.63 & 86.54 & 81.48  \\
         \midrule
         & LoRA  &90.12  & 83.25 & 84.56 & 72.55\\
       $\textsf{Deb-XL}_{r=2}$  & FFA-LoRA  & 90.28& 83.47 & 85.59 & 79.06  \\
         & RoLoRA   & 91.32 & 84.84& 86.50 & 81.34  \\
           \bottomrule
    \end{tabular}
    \caption{Results with DeBERTa-XLarge models on GLUE. Same message size in each round for three methods for each dataset. The number of clients is 3.}
    \label{tab:same_comm_deb}}
\end{table}

\subsubsection{Performance when QLoRA is Applied} In Table~\ref{tab:quantization-base-model}, we quantize the frozen pre-trained weights to 8 bit and 4 bit for each client, and apply QLoRA\cite{dettmers2023qlora}. The relative accuracy is computed as $\frac{\mathsf{Acc}_{\mathsf{FP}}-\mathsf{Avg}(\mathsf{Acc}_{8b}+\mathsf{Acc}_{4b})}{\mathsf{Acc}_{\mathsf{FP}}}$. We use sufficient finetuning parameters and the selected layer set is $\mathcal{P}_1$ to study the effect of the quantized foundation model in the federated setting.
\begin{table}[h]
{\footnotesize
    \centering
    \begin{tabular}{ccccccccccc}
    \toprule
      Methods   &  & QQP &  &  &QNLI &  &  & MNLI & & \\
         \midrule
         & FP & 8-bit & 4-bit & FP &8-bit  &4-bit  & FP &8-bit  &4-bit & Relative Acc.\\
         \midrule
     LoRA   &85.86 & 85.78 & 85.47 & 92.87 & 92.71 &92.00  &87.69  & 86.57 &  86.49 &$\downarrow 0.7\%$\\
     FFA-LoRA  & 85.74 & 85.59 & 85.35 & 92.51 & 91.20 & 91.12 &85.75 & 84.86 & 83.19 &$\downarrow 1.3\%$\\
     RoLoRA  & 85.64 & 85.51 & 85.47 & 92.48 & 92.51 & 91.63 & 87.95 &87.27 &  86.12& $\downarrow 0.7\%$\\
     \bottomrule
    \end{tabular}
    \caption{Results of RoBERTa-Large models on QQP, QNLI, MNLI with the quantized base model. The rank used is 2.}
    \label{tab:quantization-base-model}}
\end{table}

\end{document}